\documentclass[runningheads]{llncs}
\usepackage[T1]{fontenc}
\usepackage{graphicx,verbatim}
\graphicspath{{figures/assets}}
\usepackage{amsmath}
\usepackage{amssymb}
\usepackage{booktabs}
\usepackage{color}
\usepackage{multirow}
\usepackage{enumitem}
\usepackage[dvipsnames, table, xcdraw]{xcolor}
\usepackage[ruled, vlined,noend, linesnumbered, commentsnumbered]{algorithm2e}
\usepackage{tikz}
\usepackage[normalem]{ulem}
\newcommand{\mysection}[1]{\noindent\textbf{#1}}
\useunder{\uline}{\ul}{}

\begin{document}
%
\title{LUTSeg: A Longitudinal Multi-Expert Dataset for Ulcer Tissue Segmentation}


\titlerunning{LUTSeg}
%

\author{Karen Sanchez\inst{1}\and
Carlos Hinojosa\inst{1}\and
Albert A. Ávila\inst{2}\and
Andrea C. Riano-Rojas\inst{3}\and
Diego H. Romero\inst{2}\and
Jenny C. Páez\inst{2}\and
Martina Llinás\inst{2}\and
Bernard Ghanem\inst{1}
}

\authorrunning{K. Sanchez et al.}
%
\institute{King Abdullah University of Science and Technology (KAUST), Saudi Arabia \and
Subred Norte E.S.E, Hospital Simón Bolívar, Colombia\and
Universidad del Rosario, Bogotá, Colombia\\
\email{karen.sanchez@kaust.edu.sa}}

\maketitle 

\begin{abstract}

Quantifying wound tissue composition is essential for monitoring chronic ulcer progression and guiding treatment decisions. However, pixel-level annotations are costly, and multi-tissue wound datasets remain scarce, particularly for neglected diseases such as leprosy. We introduce LUTSeg, a longitudinal chronic ulcer dataset comprising 141 images from 39 patients with wound masks and five tissue categories annotated by five expert clinicians, including a multi-expert gold-standard subset for inter-rater agreement analysis. To establish an initial benchmark for LUTSeg, we further propose TiSage, a semi-supervised tissue segmentation framework that integrates multi-scale semantic priors from a frozen medical vision-language model within a teacher-student architecture. We evaluate TiSage on LUTSeg and DFUTissue, showing improvements over supervised and semi-supervised baselines in most low-label settings. Code \& data: \url{https://github.com/carlosh93/TiSage}.


\end{abstract}
\section{Introduction}
\label{sec:introduction}

Effective management of chronic ulcers requires not only accurate wound boundary segmentation but also fine-grained characterization of heterogeneous tissue types within the wound. The spatial distribution of epithelial, slough, granulation, and necrotic tissues encodes clinically actionable information about inflammation, infection risk, and healing progression \cite{bowers2020chronic,gupta2021chronic,waahlstrand2025separable}. However, dense pixel-level tissue annotation is labor-intensive, costly, and inherently subjective, even among specialized wound-care experts. Therefore, high-quality multi-tissue datasets are severely scarce. This challenge is particularly pronounced in neglected diseases such as leprosy \cite{serrano2019social}, where systematic longitudinal monitoring is essential but expert annotation resources are scarce \cite{van2021psychosocial,sanchez2024co2wounds}.\\
Most existing wound analysis studies focus on wound boundary segmentation or coarse grading due to the lack of tissue-level annotations \cite{hsu2019chronic,mukherjee2017diagnostic,chairat2021non,monroy2023automated}. Kabir et al. introduced a six-class wound tissue dataset, which remains private~\cite{kabir2025woundtissue}. DFUTissue is a diabetic foot ulcer dataset with a small labeled subset and a larger unlabeled portion designed for semi-supervised learning~\cite{dhar2024dfutissue}. ComplexWoundDB covers diverse wound etiologies but remains small-scale~\cite{pereira2022complexwounddb}. For leprosy, CO2Wounds-V2~\cite{sanchez2024co2wounds} provides wound masks but lacks tissue labels.


To address these gaps, we introduce \textbf{LUTSeg} (Leprosy Ulcer Tissue Segmentation across Time), a longitudinal dataset of 141 leprosy-related ulcer images from 39 patients, with pixel-level annotations for five tissue categories (Epithelial, Slough, Granulation, Necrotic, Other), including a 46-image multi-expert gold-standard subset annotated by five clinicians for agreement analysis.

To provide an initial benchmark and demonstrate the utility of LUTSeg for label-efficient tissue segmentation under annotation scarcity, we develop \textbf{TiSage}, a semi-supervised method that integrates multi-scale semantic priors from a frozen medical vision-language model with a confidence-gated, pixel-adaptive pseudo-label refinement strategy. We evaluate TiSage on the DFUTissue and LUTSeg datasets under diverse low-label regimes. TiSage outperforms baselines, particularly in moderate annotation settings. Our contributions are as follows:
\begin{itemize}
    \item We introduce LUTSeg, a longitudinal, multi-expert, pixel-level wound tissue segmentation dataset for leprosy-related skin ulcers, an underrepresented neglected disease setting.
    \item We provide a structured annotation protocol with wound and tissue masks, covering five tissue categories. We further quantify inter-rater variability on a gold-standard subset annotated by five clinicians, highlighting the challenge of wound tissue phenotyping.
    \item To establish an initial benchmark for LUTSeg, we propose TiSage, a semi-supervised method that integrates superpixel-based semantic priors from a frozen medical vision-language model into a teacher-student framework to improve pseudo-label quality in low-label regimes.
\end{itemize}

\section{LUTSeg Dataset}
\label{sec:proposed_dataset}


Pixel-level tissue annotations in chronic wound datasets are scarce due to high annotation burden and inter-observer variability \cite{yim2025woundcarevqa}. To our knowledge, no longitudinal datasets currently provide such labels for neglected diseases like leprosy. 


\mysection{Data Acquisition.} LUTSeg contains 141 images of leprosy-related ulcers from 39 patients, acquired during routine wound care sessions over 21 months. Images were captured by medical staff using smartphone cameras, with each image corresponding to a follow-up visit. The dataset contains 3.615 $\pm$ 1.695 visits per patient. Intervals vary according to clinical scheduling and treatment plans.

LUTSeg acquisition adhered to the Declaration of Helsinki. All data were anonymized, and written informed consent was obtained from all participants. The corresponding approval was granted by the Ethics Committee of the Sanatorio de Contratación ESE hospital in Colombia, under Act 05–21.

\begin{figure}[t]
    \centering
    \includegraphics[width=\linewidth]{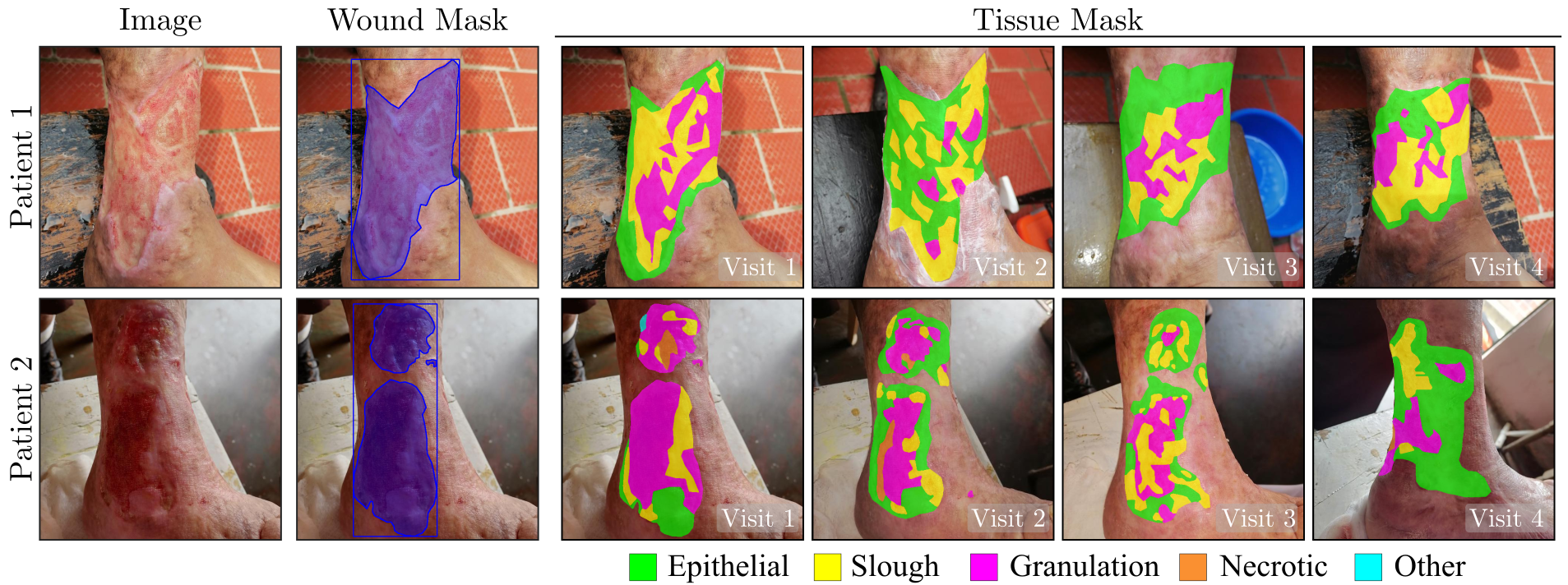}
    \caption{\small Samples of the LUTSeg dataset. First-visit image, its binary wound mask, and pixel-level segmentation into five tissue categories for each visit image.
    }
    \label{fig:dataset}
\end{figure}

\mysection{Annotation Protocol.} Pixel-level segmentation of wound tissues was performed by five specialized clinicians with expertise in complex wound care and skin tissue management. Each image was annotated independently using a standardized labeling interface. The following five tissue categories were defined in accordance with clinical practice: Epithelial, Slough, Granulation, Necrotic, and Other. Annotation was conducted in two stages: (1) Wound boundaries were delineated to produce a binary wound mask for each image. (2) All visible tissue regions within the wound area were segmented at pixel resolution into the predefined tissue categories. Given the inherent difficulty of tissue-type annotation, annotators were permitted to assign the ``Other'' category when tissue appearance did not clearly correspond to the predefined classes.

To prevent data leakage, dataset splitting was performed at the patient level, ensuring that images from the same patient were not distributed across annotation or evaluation subsets. We constructed a gold-standard subset of 46 images by selecting patients with higher tissue diversity and multiple follow-up visits. This subset comprised 46 images from 9 patients and was independently annotated by all five specialized physicians. The remaining patients were then distributed among the annotators for separate labeling. Annotation was performed using a dedicated web-based platform~\cite{LabelStudio}, ensuring standardized mask creation and quality control. Figure~\ref{fig:dataset} shows examples of our dataset with corresponding wound mask and pixel-level tissue segmentation, as well as longitudinal tissue labeling across follow-up visits. Both wound boundary masks and pixel-level tissue annotations were obtained for all follow-up visits. To derive a single reference mask per image for the gold-standard subset, we performed a consensus selection procedure. For each image, annotators voted for the most clinically accurate mask; ties were resolved via fixed-seed random selection.

\begin{figure}[t]
\includegraphics[width=\textwidth]{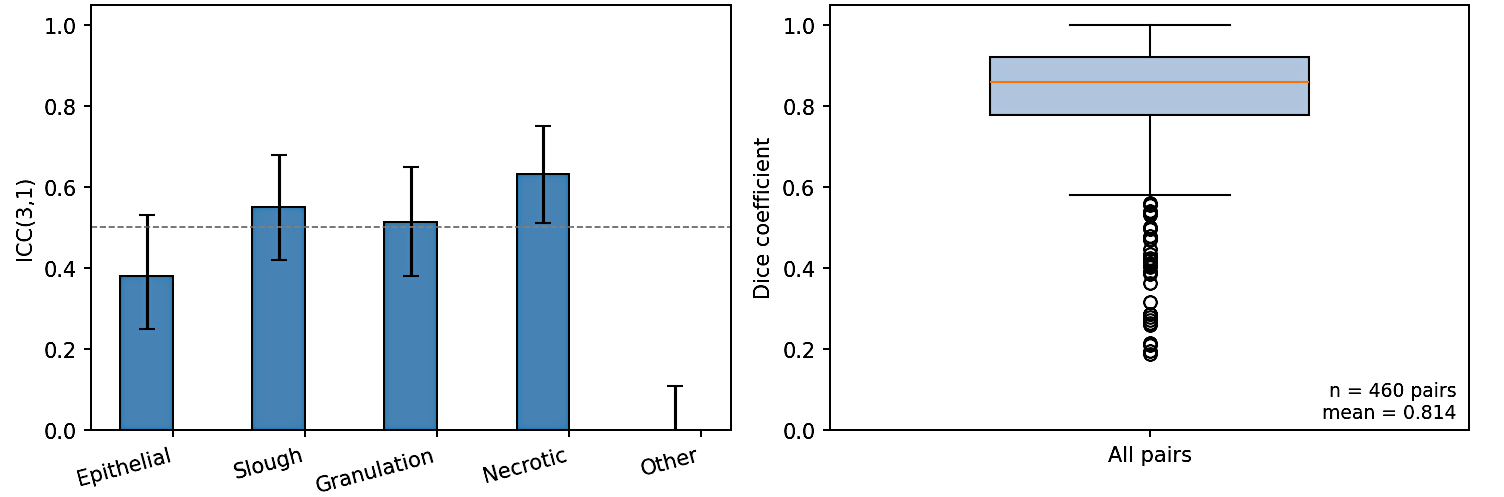}
\caption{\small (Left) ICC(3,1) for tissue proportion agreement across annotators. (Right) Pairwise Dice scores on 46 images annotated by 5 clinicians (gold-standard subset).} \label{fig:agreement}
\end{figure}


\mysection{Inter-Rater Agreement.} To quantify annotation consistency on the gold-standard subset (46 images, 5 clinicians), we evaluated inter-rater agreement using complementary metrics reflecting compositional and spatial consistency. Following the intraclass correlation framework of Shrout and Fleiss~\cite{shrout1979intraclass,liljequist2019intraclass,ramachandram2022fully}, and given that the same fixed set of raters annotated all images, we used a two-way mixed-effects model and report the single-measure ICC(3,1) (Fig.~\ref{fig:agreement}(left)). For each image and rater, we computed the proportion of each tissue type relative to the total wound area (background excluded), and ICC(3,1) was calculated independently for each tissue category. Second, we assessed spatial overlap using the Dice coefficient between all annotator pairs. Dice was computed on binary tissue-versus-background masks across all annotator pairs (460 comparisons). Agreement on tissue proportions was moderate for Necrotic (ICC = 0.63, 95\% CI 0.51-0.75), Slough (0.55, 0.42-0.68), and Granulation (0.51, 0.38-0.65), and lower for Epithelial (0.38, 0.25-0.53). The ``Other'' class showed the lowest agreement (ICC $\approx$ 0 (95\% CI -0.08-0.11)), reflecting its role in capturing visually ambiguous or heterogeneous regions. Pairwise Dice scores were high overall (mean 0.814, median 0.860; Fig.~\ref{fig:agreement}(right)), but exhibited substantial outliers (minimum 0.187), indicating disagreement in tissue extent and boundary delineation in some cases.
These findings highlight the intrinsic subjectivity of wound tissue phenotyping, particularly at ambiguous boundaries and for rare or heterogeneous tissue patterns. Even among specialized clinicians, substantial variability persists. These observations motivate TiSage, which incorporates uncertainty-aware, confidence-weighted learning to account for clinical ambiguity.

\section{TiSage Method}

\begin{figure*}[t]
    \centering
    \includegraphics[width=\linewidth]{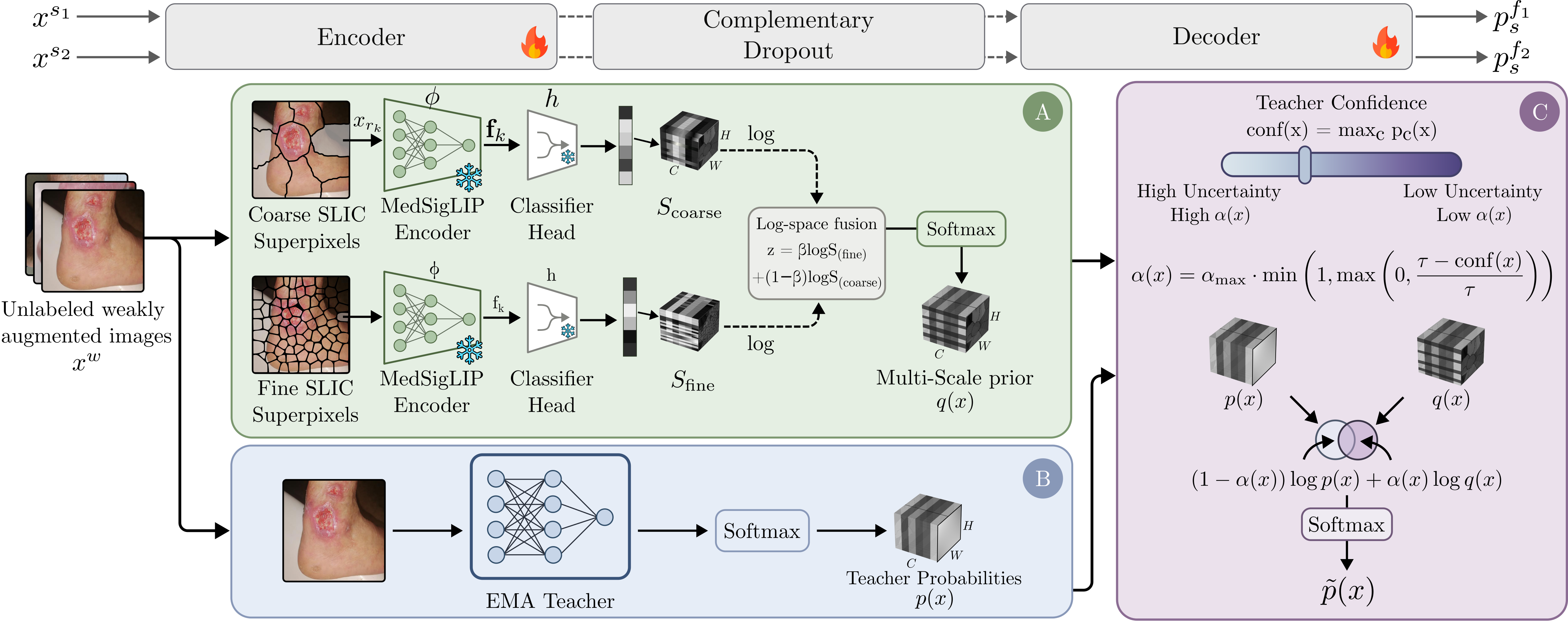}
    \caption{\small (A) Multi-scale semantic prior construction: coarse and fine SLIC superpixels are encoded with a frozen MedSigLIP encoder, classified, and fused in log space to produce a multi-scale prior $q(x)$. (B) EMA teacher predictions $p(x)$ from weakly augmented images. (C) Pixel-adaptive fusion of teacher and prior based on teacher confidence, yielding calibrated pseudo-labels $\tilde{p}(x)$.}
    \label{fig:method}
\end{figure*}

Semi-supervised wound tissue segmentation is challenging due to the scarcity of annotations and inter-observer variability. In low-label regimes, teacher–student methods such as UniMatch-V2~\cite{yang2025unimatch} rely on pseudo-labels, which may propagate errors into ambiguous regions. We propose TiSage, which integrates multi-scale semantic priors from a frozen medical vision-language model with pixel-adaptive pseudo-label calibration. Our framework consists of: (i) a MedSigLIP-based~\cite{sellergren2025medgemma} superpixel prior, (ii) multi-scale log-space fusion, and (iii) pixel-adaptive teacher–prior fusion with entropy-weighted supervision (Fig.~\ref{fig:method}).

\subsection{Multi-Scale Semantic Prior Construction}



\mysection{Semantic Prior.} Let $\mathcal{D}_L = \{(x_i, y_i)\}_{i=1}^{N_L}$ denote the labeled set and $\mathcal{D}_U = \{x_j\}_{j=1}^{N_U}$ the unlabeled set. Given an image $x$, we generate superpixels using SLIC~\cite{achanta2010slic}. For each region $r_k$, we pad it to a square and resize it to $448 \times 448$ before feeding it to MedSigLIP. We compute a normalized embedding $\mathbf{f}_k = \frac{\phi(x_{r_k})}{\|\phi(x_{r_k})\|_2}$, where $\phi(\cdot)$ denotes the MedSigLIP encoder. To train the region-level classifier, we apply SLIC to labeled images $(x_i, y_i) \in \mathcal{D}_L$ and assign each superpixel a class via majority voting over ground-truth pixels within the region. A lightweight linear classification head $h$ is trained once on the region embeddings $\mathbf{f}_k$ using class-balanced cross-entropy and kept frozen during semi-supervised segmentation training. Region-level logits and probabilities are $\boldsymbol{\ell}_k = h(\mathbf{f}_k)$ and $\boldsymbol{\pi}_k = \mathrm{softmax}(\boldsymbol{\ell}_k)$. We broadcast $\boldsymbol{\pi}_k$ to every pixel in $r_k$ to obtain a dense per-pixel semantic prior:
\begin{equation}
		S(x) \in [0,1]^{C \times H \times W}, \quad \sum_{c=0}^{C-1} S_c(x;u,v) = 1 \;\;\forall (u,v).
\end{equation}



\mysection{Multi-Scale Fusion.} Single-scale superpixels involve a trade-off between spatial smoothness (coarse regions) and boundary precision (fine regions). To balance these effects, we compute two priors: coarse prior $S_{\text{coarse}}$ and fine prior $S_{\text{fine}}$. We fuse them in log-probability space:
\begin{equation}
		z = \beta \log S_{\text{fine}} + (1-\beta) \log S_{\text{coarse}}; \quad q = \text{softmax}(z),
        \label{eq:multi-scale-fusion}
\end{equation}
where $\beta \in [0,1]$ controls the balance between fine and coarse scales. 


\subsection{Pixel-Adaptive Teacher--Prior Fusion}

Let $p(x)$ denote the pixel-wise class probabilities predicted by the EMA teacher on weakly augmented unlabeled images, and let $q(x)$ denote the multi-scale semantic prior computed on the same view (Eq. \eqref{eq:multi-scale-fusion}). We employ a pixel-adaptive fusion that modulates the influence of the prior according to teacher confidence $\text{conf}(x) = \max_c p_c(x)$. Hence, we define:
\begin{equation}
    \alpha(x) = \alpha_{\max} \cdot \min \left(1, \max \left(0, \frac{\tau - \mathrm{conf}(x)}{\tau}\right)
		\right),
\end{equation}
where $\tau$ is a confidence threshold and $\alpha_{\max}\in [0,1]$. Fusion is performed in log-probability space, corresponding to a weighted product-of-experts:
\begin{equation}
\tilde{p}(x) = \text{softmax}\left(
(1-\alpha(x)) \log p(x) + \alpha(x) \log q(x)
\right).
\end{equation}
Thus, the prior has greater influence when the teacher is uncertain and negligible influence when the teacher is confident.

\subsection{Training Objectives}


For labeled data, we use the standard cross-entropy loss $\mathcal{L}_{\text{sup}} = \text{CE}(y, p_s)$, where $p_s$ is the student prediction. We denote the weakly and strongly augmented views of an image $x$ as $x^w$ and $x^s$, respectively. For unlabeled data, we compute pseudo-labels from the weak view and apply them to the strongly augmented views, following UniMatch-V2~\cite{yang2025unimatch}. We supervise the student using calibrated pseudo-labels $\tilde{p}(x)$ via a blend of hard and soft supervision:
\begin{equation}
		\mathcal{L}_{\text{unsup}} =
		(1 - \beta_{\text{soft}}) \mathcal{L}_{\text{CE}}^{\text{hard}}
		+ \beta_{\text{soft}} \mathcal{L}_{\text{KL}}^{\text{soft}},
\end{equation}
where we set $\beta_{\text{soft}}=0.5$ unless otherwise noted. The hard term uses $\hat{y}(x)=\arg\max_c \tilde{p}_c(x)$ and follows UniMatch-V2 by masking pixels with low pseudo-label confidence ($\max_c \tilde{p}_c(x) < \gamma$):
\begin{equation}
		\mathcal{L}_{\text{CE}}^{\text{hard}} = \sum_x m(x)\,\text{CE}\big(\hat{y}(x), p_s(x)\big), \quad
		m(x)=\mathbf{1}\!\left[\max_c \tilde{p}_c(x) \ge \gamma\right].
\end{equation}
To avoid relying on hard thresholding for soft supervision, we weight the soft KL term by entropy:
\begin{equation}
		w(x) = 1 - \frac{H(\tilde{p}(x))}{\log C}; \quad \mathcal{L}_{\text{KL}}^{\text{soft}}
		= \sum_x w(x) \, \text{KL}\big(\tilde{p}(x) \| p_s(x)\big),
\end{equation}
where $H(\cdot)$ denotes Shannon entropy. The overall loss is $\mathcal{L}=\mathcal{L}_{\text{sup}}+\mathcal{L}_{\text{unsup}}$.


\section{Experiments}
\label{sec:experiments}

\mysection{Datasets.} We evaluate TiSage on both a public benchmark (DFUTissue) and our proposed dataset. DFUTissue~\cite{dhar2024dfutissue} is a publicly available dataset of diabetic foot ulcer images with pixel-level tissue annotations. It includes wound masks and four tissue categories (granulation, slough, eschar/necrotic, and epithelial). Following prior work, we adopt the standard splits and low-label regimes.

\mysection{Metrics.} Performance is measured using mean Intersection-over-Union (mIoU) and Dice coefficient. Unless otherwise stated, reported values correspond to the EMA teacher model at inference, following prior works \cite{sohn2020fixmatch,yang2025unimatch}. 

\mysection{Implementation Details.} TiSage is built upon the UniMatch-V2~\cite{yang2025unimatch} teacher– student framework. MedSigLIP is used as a frozen encoder to extract superpixel-level embeddings. Low-label regimes are simulated using 1/4, 1/8, and 1/16 labeled data splits. All methods use the same backbone and training schedule. We report mean performance over three random seeds (0, 1, and 2).
\subsection{Quantitative Results}

\begin{table}[t]
\caption{\small Comparison of supervised and semi-supervised segmentation methods. In DFUTissue, \textbf{Fixed} denotes the official predefined SSL split. Supervised methods use only the labeled portion of each split, whereas semi-supervised methods additionally use the corresponding unlabeled images. DeepLabV3+ uses an ImageNet-pretrained ResNet-50 encoder; DINOv2-DPT and all SSL methods use DINOv2-Base. For LUTSeg, the fully supervised references using all 111 labeled training images are $31.37/39.19$ mIoU/Dice for DINOv2--DPT and $22.06/25.61$ for DeepLabV3+.}
\label{tab:baseline_tisage}
\resizebox{\linewidth}{!}{%
\begin{tabular}{lcccccccc|cccccc}
\hline
                         & \multicolumn{8}{c|}{DFUTissue}                                                                                                  & \multicolumn{6}{c}{LUTSeg}                                                         \\ \cline{2-15} 
                         & \multicolumn{2}{c}{Fixed}       & \multicolumn{2}{c}{1/4}         & \multicolumn{2}{c}{1/8}         & \multicolumn{2}{c|}{1/16} & \multicolumn{2}{c}{1/4}         & \multicolumn{2}{c}{1/8}         & \multicolumn{2}{c}{1/16}        \\ \cline{2-15} 
\multirow{-3}{*}{Method} & mIoU           & F1             & mIoU           & Dice           & mIoU           & Dice           & mIoU        & Dice        & mIoU           & Dice           & mIoU           & Dice           & mIoU           & Dice           \\ \hline
\multicolumn{15}{l}{\textit{Supervised methods}} \\
DeepLabV3+--R50           & 70.02          & 81.12          & 65.22          & 77.14          & 58.32          & 70.52          & 48.38       & 58.87       & 19.89          & 23.17          & 20.79          & 24.18          & 21.25          & 25.34          \\
DINOv2--DPT        & 68.71          & 80.21          & 66.23          & 78.03          & 64.68          & 76.57          & 52.83       & 65.02       & 28.38          & 35.19          & 20.47          & 23.55          & 24.47          & 30.15          \\ \hline
\multicolumn{15}{l}{\textit{Semi-supervised methods}} \\
FixMatch                 & 68.91          & 80.19          & 67.17          & 78.80          & 66.90          & 78.40          & 60.14       & 71.30       & 27.70          & 33.00          & 27.26          & 33.91          & 27.42          & 34.33          \\
UniMatch-V2              & 69.94          & 80.96          & 68.17          & 79.67          & 67.28          & 78.85          & 61.80       & 73.24       & 26.13          & 30.55          & 27.60          & 34.24          & 27.35          & 32.24          \\
\rowcolor[HTML]{EFEFEF} 
TiSage (Ours)            & \textbf{72.36} & \textbf{83.05} & \textbf{69.77} & \textbf{81.00} & \textbf{67.93} & \textbf{79.28} & 61.33       & 73.17       & \textbf{28.73} & \textbf{34.50} & \textbf{31.70} & \textbf{39.25} & \textbf{28.55} & \textbf{34.04} \\ \hline
\end{tabular}%
}
\end{table}


\mysection{Comparison with baselines.} We compare TiSage against two supervised baselines, DINOv2--DPT~\cite{oquabdinov2,ranftl2021vision} and DeepLabV3+--R50~\cite{chen2018encoder}, and two semi-supervised baselines, FixMatch~\cite{sohn2020fixmatch} and UniMatch-V2~\cite{yang2025unimatch}. Table~\ref{tab:baseline_tisage} reports segmentation performance under different labeled regimes. Supervised methods are trained solely on the labeled portion of each split (no unlabeled data or pseudo-labeling). TiSage consistently outperforms UniMatch-V2 in six of seven settings, with notable gains on DFUTissue Fixed and 1/4 (+2.42 and +1.60 mIoU) and LUTSeg 1/8 (+4.10 mIoU); at DFUTissue 1/16, TiSage remains competitive, trailing by only 0.47 mIoU. TiSage also surpasses supervised approaches across all splits. 

\begin{table}[t]
\centering
\caption{\small Per-class IoU (\%) results on the DFUTissue and LUTSeg for 1/8 label regime.}\label{tab:perclass_iou}
\resizebox{0.95\linewidth}{!}{%
\begin{tabular}{lccccc|ccccccc}
\hline
\multirow{2}{*}{Method} & \multicolumn{5}{c|}{DFUTissue}     & \multicolumn{7}{c}{LUTSeg}                       \\ \cline{2-13} 
                        & Bg   & Fibrin & Gran. & Callus & mIoU  & Bg   & Epi  & Slough & Gran. & Necr. & Other & mIoU  \\ \hline
UniMatch-V2             & 88.3 & 47.3   & 86.9  & 57.2   & 69.94 & 94.8 & 25.4 & 5.6    & 39.7  & 0.0   & 0.0   & 27.60 \\
TiSage (Ours)           & 88.3 & 57.6   & 86.9  & 56.6   & 72.36 & 95.7 & 26.5 & 15.4   & 52.4  & 0.1   & 0.0   & 31.70 \\ \hline
\end{tabular}%
}
\vspace{-0.5em}
\end{table}

\mysection{Per-class IoU.} Table~\ref{tab:perclass_iou} reports per-class IoU under the 1/8 labeled regime. On DFUTissue, TiSage notably improves Fibrin (+10.3 IoU), a clinically ambiguous category, while maintaining performance on dominant classes. On LUTSeg, gains are more pronounced, particularly for Slough (+9.8 IoU) and Granulation (+12.7 IoU), which exhibit higher variability and lower baseline performance. These results indicate that multi-scale semantic guidance primarily benefits minority and ambiguous tissue classes under annotation scarcity.

\begin{table}[t]
\centering
\caption{\small MedSigLIP prior-only performance on DFUTissue and LUTSeg on val split.}
\label{tab:prior_ceiling}
\resizebox{0.86\linewidth}{!}{%
\begin{tabular}{ccc|cc}
\hline
\multirow{2}{*}{Prior setup}     & \multicolumn{2}{c|}{DFUTissue} & \multicolumn{2}{c}{LUTSeg} \\
                                 & Pixel Acc (\%)   & mIoU (\%)   & Pixel Acc (\%) & mIoU (\%) \\ \hline
Zero-shot (single-scale SLIC)    & 76.24            & 34.70       & 67.94          & 16.39     \\
Classifier (single-scale SLIC)   & 79.40            & 45.67       & 80.06          & 24.02     \\
Classifier (coarse only)         & 74.71            & 44.70       & 77.48          & 23.38     \\
Classifier (fine only)           & 81.55            & 48.05       & 81.00          & 25.44     \\
Classifier (fused) & 80.92            & 52.11       & 81.77          & 26.28     \\ \hline
\end{tabular}%
}
\end{table}

\mysection{Ablation Studies.} To assess the standalone capability of the MedSigLIP prior, we evaluate prior-only segmentation without the encoder-decoder architecture of TiSage (Table~\ref{tab:prior_ceiling}). While the prior alone is insufficient for high-quality segmentation, multi-scale fusion substantially improves mIoU over single-scale variants, validating the proposed log-space fusion strategy. Component ablations on DFUTissue (1/8 split) confirm that each TiSage module contributes to performance, with entropy-weighted KL having the largest impact (-0.60 mIoU when removed). Sensitivity analysis on LUTSeg (1/8 split) shows that performance varies by $\leq 0.35$ mIoU across $\tau \in [0.80, 0.95]$, with the best result at $\alpha_{\max}=0.25$.

\subsection{Qualitative Results}

\begin{figure}[bt]
    \centering
    \includegraphics[width=\linewidth]{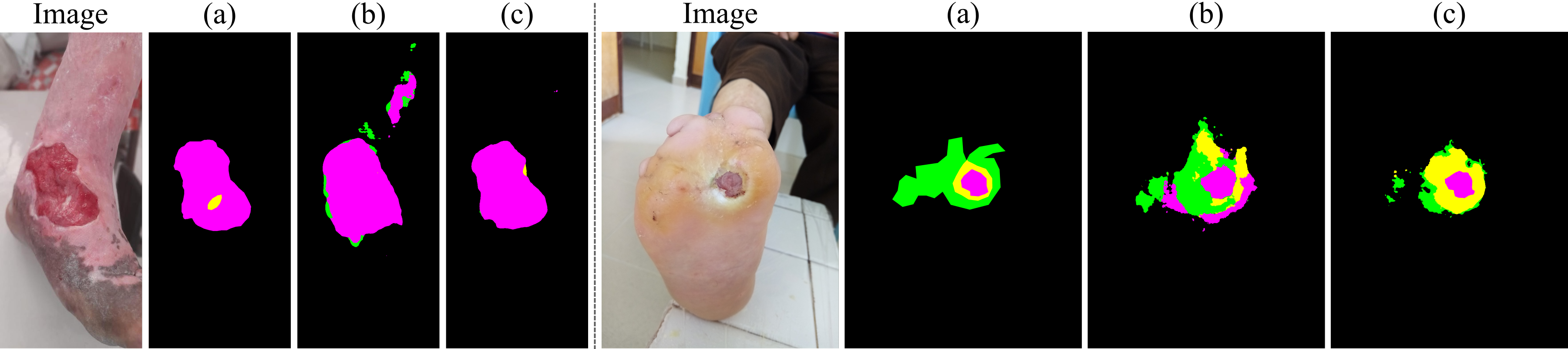}
    \caption{\small Qualitative comparison of (a) Ground Truth, (b) UniMatch-V2, and (c) TiSage results for two random samples from the LUTSeg proposed dataset.
    }
    \label{fig:visual_res}
    \vspace{-0.5em}
\end{figure}

\noindent Figure~\ref{fig:visual_res} shows qualitative results on LUTSeg. Compared to UniMatch-V2, TiSage produces more coherent boundaries and fewer fragmented predictions in ambiguous regions, reflecting the effect of our pixel-adaptive semantic fusion.

%


\section{Conclusions}
\label{sec:conclusion}
We introduced LUTSeg, a longitudinal, multi-expert, pixel-level wound tissue segmentation dataset for ulcers caused by a neglected tropical disease.
We further proposed TiSage, a semi-supervised framework that leverages semantic guidance to improve robustness under annotation scarcity. Experiments on LUTSeg and DFUTissue show consistent gains over established baselines. Together, they set a benchmark for efficient tissue segmentation under realistic clinical constraints.

\begin{credits}
\subsubsection{\ackname}
The research reported in this publication was supported by funding from King Abdullah University of Science and Technology (KAUST) - Center of Excellence for Generative AI, under award number 5940.
\subsubsection{\discintname}
Authors declare that they have no conflict of interest.
\end{credits}

\bibliographystyle{splncs04}
\bibliography{mybibliography}
%




\end{document}